\documentclass[10pt,twocolumn,letterpaper]{article}

\usepackage{cvpr}
\usepackage{times}
\usepackage{epsfig}
\usepackage{graphicx}
\usepackage{amsmath}
\usepackage{amssymb}
\usepackage{algorithm}
\usepackage[noend]{algpseudocode}
\usepackage{bm}

\usepackage[breaklinks=true,bookmarks=false]{hyperref}
\hypersetup{colorlinks}

\cvprfinalcopy 

\newcommand{\etal}[0]{et al.}
\newcommand{\boldhead}[1]{\vspace{0.03in}\noindent\textbf{#1: }}

\begin{document}

\title{LayoutNet: Reconstructing the 3D Room Layout from a Single RGB Image}

\author{Chuhang Zou$^\dagger$\\
\and
Alex Colburn$^\ddagger$\\
\and
Qi Shan$^\ddagger$\\
\and
Derek Hoiem$^\dagger$
\and
$^\dagger$University of Illinois at Urbana-Champaign\\
{\tt\small \{czou4, dhoiem\}@illinois.edu}
\and
$^\ddagger$Zillow Group\\
{\tt\small \{alexco, qis\}@zillow.com}
}

\maketitle

\begin{abstract}
We propose an algorithm to predict room layout from a single image that generalizes across panoramas and perspective images, cuboid layouts and more general layouts (e.g. ``L"-shape room). 
Our method operates directly on the panoramic image, rather than decomposing into perspective images as do recent works.  Our network architecture is similar to that of RoomNet~\cite{lee2017roomnet}, but we show improvements due to aligning the image based on vanishing points, predicting multiple layout elements (corners, boundaries, size and translation), and fitting a constrained Manhattan layout to the resulting predictions.  
Our method compares well in speed and accuracy to other existing work on panoramas, achieves among the best accuracy for perspective images, and can handle both cuboid-shaped and more general Manhattan layouts.
\end{abstract}

\section{Introduction}
Estimating the 3D layout of a room from one image is an important goal, with applications such as robotics and virtual/augmented reality.  The room layout specifies the positions, orientations, and heights of the walls, relative to the camera center. The layout can be represented as a set of projected corner positions or boundaries, or as a 3D mesh. Existing works apply to special cases of the problem, such as predicting cuboid-shaped layouts from perspective images or from panoramic images.

\begin{figure}[ht]
\begin{center}
   \includegraphics[width=1.0\linewidth]{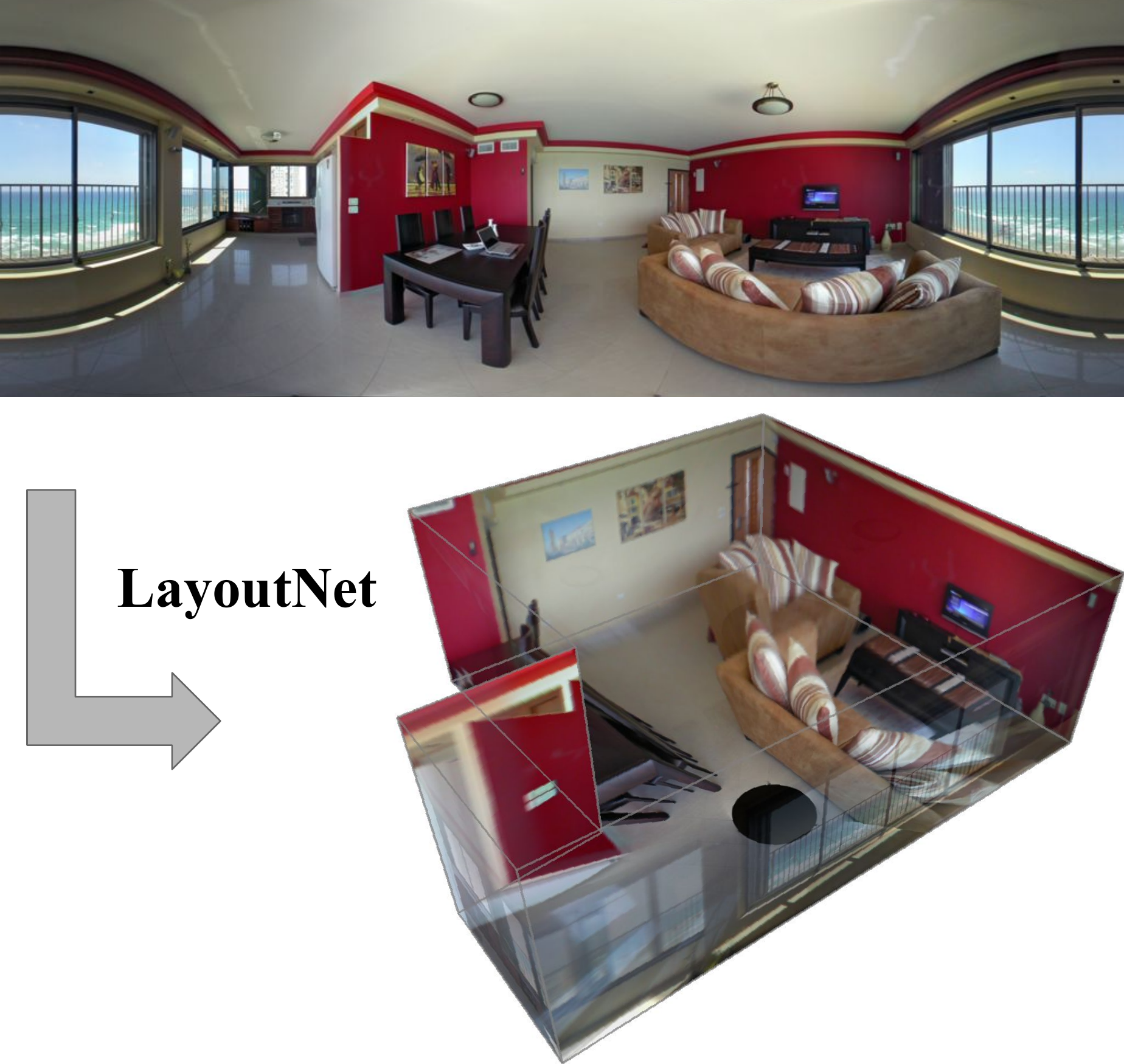}
\end{center}
\vspace{-1.0em}
   \caption{\textbf{Illustration.} Our LayoutNet predicts a non-cuboid room layout from a single panorama under equirectangular projection.}
 \vspace{-1.0em}
\label{fig:illustration}
\end{figure}

We present LayoutNet, a deep convolution neural network~(CNN) that estimates the 3D layout of an indoor scene from a single perspective or panoramic image~(Figure.~\ref{fig:illustration}). 
Our method compares well in speed and accuracy on panoramas and is among the best on perspective images. Our method also generalizes to non-cuboid Manhattan layouts, such as ``L''-shaped rooms. 
Code is available at:~\url{https://github.com/zouchuhang/LayoutNet}.


\begin{figure*}[ht]
\begin{center}
\includegraphics[width=1.0\linewidth]{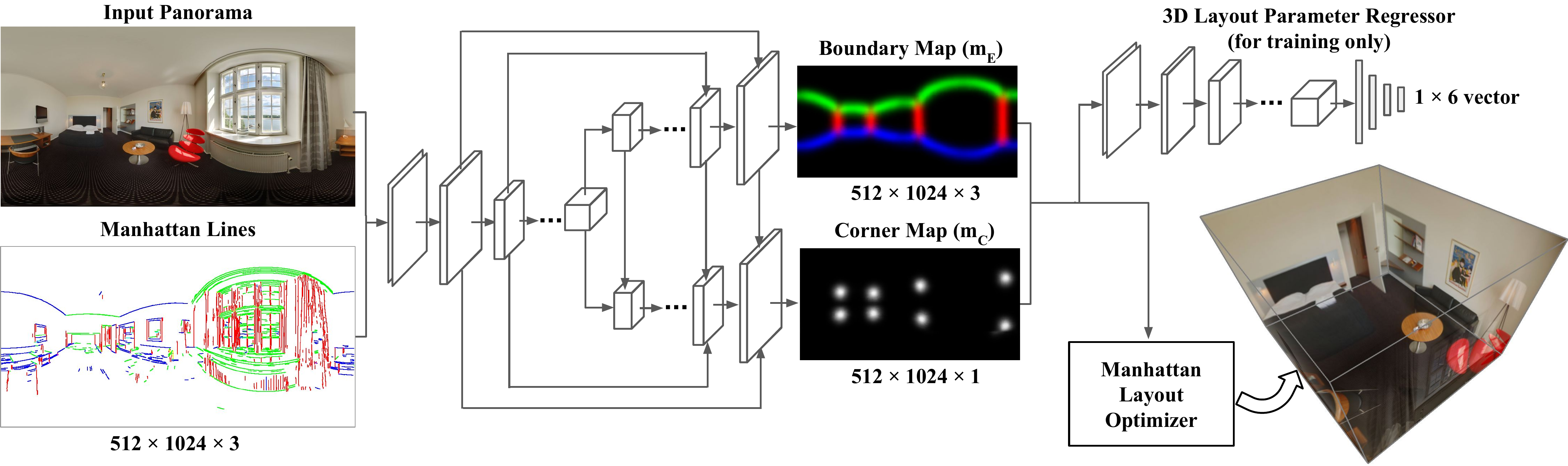}
\end{center}
\vspace{-3mm}
   \caption{\textbf{Overview.} Our LayoutNet follows the encoder-decoder strategy. The network input is a concatenation of a single RGB panorama and Manhattan line map. The network 
   jointly predicts layout boundaries and corner positions. The 3D layout parameter loss 
   encourages predictions that maximize accuracy. The final prediction is a Manhattan constrained layout reconstruction. Best viewed in color.}
   \vspace{-1.0em}
\label{fig:overview}
\end{figure*}

Our LayoutNet approach operates in three steps~( Figure.~\ref{fig:overview}). First, our system analyzes the vanishing points and aligns the image to be level with the floor (Sec.~\ref{text:preprocess}). This alignment ensures that wall-wall boundaries are vertical lines and substantially reduces error according to our experiments. In the second step, corner (layout junctions) and boundary probability maps are predicted directly on the image using a CNN with an encoder-decoder structure and skip connections (Sec.~\ref{text:network}). Corners and boundaries each provide a complete representation of room layout. We find that jointly predicting them in a single network leads to better estimation. Finally, the 3D layout parameters are optimized to fit the predicted corners and boundaries (Sec.~\ref{text:fitting}). The final 3D layout loss from our optimization process is difficult to back-propagate through the network, but direct regression of the 3D parameters during training serves as an effective substitute, encouraging predictions that maximize accuracy of the end result. 


Our contributions are:
\begin{itemize}
\vspace{-0.7em}
    \item We propose a more general RGB image to layout algorithm that is suitable for perspective and panoramic images with Manhattan layouts. Our system 
    compares well in speed and accuracy for panoramic images and achieves the second best for perspective images, while also being the fastest.
    \vspace{-0.9em}
    \item We demonstrate gains from using precomputed vanishing point cues, geometric constraints, and post-process optimization, indicating that deep network approaches still benefit from explicit geometric cues and constraints.  We also show that adding an objective to directly regress 3D layout parameters leads to better predictions of the boundaries and corners that are used to solve for the final predicted layout.
    \vspace{-0.9em}
    \item We extend the annotations for the Stanford {2D-3D} dataset~\cite{armeni2017joint}, providing room layout annotations that can be used in future work.  
\end{itemize}

\section{Related Work}

Single-view room layout estimation has been an active topic of research for the past ten years. Delage \etal~\cite{delage2006dynamic} fit floor/wall boundaries in a perspective image taken by a level camera to create a 3D model under ``Manhattan world'' assumptions~\cite{coughlan1999manhattan}. The Manhattan world assumptions are that all walls are at right angles to each other and perpendicular to the floor. A special case is the cuboid model, in which four walls, ceiling, and floor enclose the room.  Lee \etal~\cite{lee2009geometric} produce Orientation Maps, generate layout hypotheses based on detected line segments, and select a best-fitting layout from among them. Hedau \etal~\cite{hedau2009recovering} recover cuboid layouts by solving for three vanishing points, sampling layouts consistent with those vanishing points, and selecting the best layout based on edge and Geometric Context~\cite{hoiem2005geometric} consistencies. 
Subsequent works follow a similar approach, with improvements to layout generation~\cite{schwing2012efficient_cvpr,schwing2012efficient_eccv,ramalingam2013manhattan}, features for scoring layouts~\cite{schwing2012efficient_eccv,ramalingam2013manhattan}, and incorporation of object hypotheses~\cite{hedau2010thinking,gupta2010estimating,del2012bayesian,del2013understanding,zhao2013scene} or other context.  The most recent methods train deep network features to classify pixels into layout surfaces (walls, floor, ceiling)~\cite{dasgupta2016delay,izadinia2017im2cad}, boundaries~\cite{mallya2015learning}, corners~\cite{lee2017roomnet}, or a combination~\cite{CFILE16}. 

Nearly all of these works aim to produce cuboid-shaped layouts from perspective RGB images. A few works also operate on panoramic images. Zhang~\etal~\cite{zhang2014panocontext} propose the PanoContext dataset and method to estimate room layout from 360$^\circ$ panoramic images (more on this later). Yang~\etal~\cite{yang2016efficient} recover layouts from panoramas based on edge cues, Geometric Context, and other priors. Xu~\etal~\cite{Pano2CAD} estimate layout based on surface orientation estimates and object hypotheses. Other works recover indoor layout from multiple images (e.g.,~\cite{cabral2014piecewise}) or RGBD images (e.g.,~\cite{DerekRGBD2012,zhang2013estimating,guo2015predicting,liu2016layered}), where estimates rely heavily on 3D points obtained from sensors or multiview constraints. Rent3D~\cite{liu2015rent3d} takes advantage of a known floor plan. 
Our approach 
simplifies reconstruction by estimating layout directly on a single RGB equirectangular panorama. 
Our final output is a sparse and compact planar Manhattan layout parameterized by each wall's distance to camera, height, and the layout rotation. 

Our work is most similar in goal to PanoContext~\cite{zhang2014panocontext} and in approach to RoomNet~\cite{lee2017roomnet}. PanoContext extends the frameworks designed for perspective images to panoramas, estimating vanishing points, generating hypotheses, and scoring hypotheses according to Orientation Maps, Geometric Context, and object hypotheses. To compute these features, PanoContext first projects the panoramic image into multiple overlapping perspective images, and then combines the feature maps back into a panoramic image. Our approach is more direct: after aligning the panoramic image based on vanishing points, our system uses a deep network to predict boundaries and corners directly on the panoramic image.  In this regard, we are similar to RoomNet, which uses a deep network to directly predict layout corners in perspective images, as well as a label that indicates which corners are visible. Our method differs from RoomNet in several ways. Our method applies to panoramic images.  Our method also differs in the alignment step (RoomNet performs none) and in our multitask prediction of boundaries, corners, and 3D cuboid parameters.  Our final inference is constrained to produce a Manhattan 3D layout. RoomNet uses an RNN to refine 2D corner position predictions, but those predictions might not be consistent with any 3D cuboid layout. 
Our experiments show that all of these differences improve results.

More generally, we propose the first method, to our knowledge, that applies to both perspective and panoramic images. 
We also show that our method extends easily to non-cuboid Manhattan layouts. Thus, our method is arguably the most general and effective approach to date for indoor layout estimation from a single RGB image.

\section{Approach}\label{text:approach}

We first describe our method for predicting cuboid-shaped layouts from panoramas: alignment (Sec.~\ref{text:preprocess}), corner and boundary prediction with a CNN (Sec.~\ref{text:network} and~\ref{text:training}), and optimization of 3D cuboid parameters (Sec.~\ref{text:fitting}). Then, we describe modifications to predict on more general (non-cuboid) Manhattan layouts and perspective images~(Sec.~\ref{text:exten}).

\subsection{Panoramic image alignment}\label{text:preprocess}
Given the input as a panorama that covers a $360^\circ$ horizontal field of view, we first align the image by estimating the floor plane direction under spherical projection, rotate the scene, and reproject it to the 2D equirectangular projection. Similar to Zhang \etal's approach~\cite{zhang2014panocontext}, we select long line segments using the Line Segment Detector (LSD)~\cite{randall2012_lsd} in each overlapped perspective view, then vote for three mutually orthogonal vanishing directions using the Hough Transform. This pre-processing step eases our network training. The detected candidate Manhattan line segments also provide additional input features that improve the performance, as shown in Sec.~\ref{text:experiments}. 

\subsection{Network structure} \label{text:network}
An overview of the LayoutNet network is illustrated in Fig.~\ref{fig:overview}. The network follows an encoder-decoder strategy. 

\boldhead{Deep panorama encoder} The input is a 6-channel feature map: the concatenation of single RGB panorama with resolution of $512\times 1024$ (or $512\times 512$ for perspective images) and the Manhattan line feature map lying on three orthogonal vanishing directions using the alignment method in Sec.~\ref{text:preprocess}. The encoder contains $7$ convolution layers with kernel size of $3\times 3$. Each convolution is followed by a ReLU operation and a max pooling layer with the down-sampling factor of $2$. The first convolution contains 32 features, and we double size after each convolution. 
This deep structure ensures a better feature learning from high resolution images and help ease the decoding step. We tried Batch Normalization after each convolution layer 
but observe lower accuracy. 
We also explored an alternative structure that applies a separate encoder for the input image and the Manhattan lines, but observe no increase in performance compared to our current simpler design.

\boldhead{2D layout decoder} The decoder consists of two branches as shown in Fig.~\ref{fig:overview}. The top branch, the layout boundary map~($\bm{m_E}$) predictor, decodes the bottleneck feature into the 2D feature map with the same resolution as the input. $\bm{m_E}$ is a 3-channel probability prediction of wall-wall, ceiling-wall and wall-floor boundary on the panorama, for both visible and occluded boundaries. The boundary predictor contains $7$ layers of nearest neighbor up-sampling operation, each followed by a convolution layer with kernel size of $3\times 3$, and the feature size is halved through layers from $2048$. The final layer is a Sigmoid operation. We add skip connections to each convolution layer following the spirit of the U-Net structure~\cite{ronneberger2015_unet}, in order to prevent shifting of predictions results from the up-sampling step. The lower branch, the 2D layout corner map~($\bm{m_C}$) predictor, follows the same structure as the boundary map predictor and additionally receives skip connections from the top branch for each convolution layer. This stems from the intuition that layout boundaries imply corner positions, especially for the case when a corner is occluded. We show in our experiments~(Sec.~\ref{text:experiments}) that the joint prediction helps improve the accuracy of the both maps, leading to a better 3D reconstruction result. We experimented with fully convolutional layers~\cite{long_shelhamer_fcn} instead of the up-sampling plus convolutions structure, but observed worse performance with checkerboard artifacts.



\boldhead{3D layout regressor} The function to map from 2D corners and boundaries to 3D layout parameters is simple mathematically, but difficult to learn. So we train a regressor for 3D layout parameters with the purpose of producing better corners and boundaries, rather than for its own sake. As shown in Fig.~\ref{fig:overview}, the 3D regressor gets as input the concatenation of the two predicted 2D maps and predicts the parameters of the 3D layout. We parameterize the layout with 6 parameters, assuming the ground plane is aligned on the $x-z$ axis: width $s_w$, length $s_l$, height $s_h$, translation $T = (t_x,t_z)$ and rotation $r_\theta$ on the $x-z$ plane. The regressor follows an encoder structure with $7$ layers of convolution with kernel size $3\times 3$, each followed by a ReLU operation and a max pooling layer with the down sampling factor of 2. The convolution feature size doubles through layers from the input 4 feature channel. The next four fully-connected layers have sizes of 1024, 256, 64, and 6, with ReLU in between. The output $1\times 6$ feature vector $d = \{s_w, s_l, s_h, t_x, t_z, r_\theta\}$ is our predicted 3D cuboid parameter. Note that the regressor outputs the parameters of the 3D layout that can be projected back to the 2D image, presenting an end-to-end prediction approach. We observed that the 3D regressor is not accurate~(with corner error of 3.36\% in the PanoContext dataset compared with other results in Table~\ref{tab:cuboid}), but including it in the loss objective tends to slightly improve the predictions of the network. The direct 3D regressor fails due to the fact that small position shifts in 2D can have a large difference in the 3D shape, making the network hard to train.

\textbf{Loss function.} The overall loss function of the network is in Eq.~\ref{equ:loss}:
\vspace{-1em}
\begin{align}\label{equ:loss}
    L(\bm{m_E},\bm{m_C},\bm{d}) &= -\alpha \frac{1}{n}\sum_{p\in \bm{m_E}}\big(\hat{p}\log p +(1-\hat{p})\log (1-p) \big)\nonumber \\
    &-\beta \frac{1}{n}\sum_{q\in \bm{m_C}}\big(\hat{q}\log q +(1-\hat{q})\log (1-q) \big)\nonumber \\
    &+ \tau \|\bm{d}-\bm{\hat{d}}\|_2
\end{align}
The loss is the summation over the binary cross entropy error of the predicted pixel probability in $\bm{m_E}$ and $\bm{m_C}$ compared to ground truth, plus the Euclidean distance of regressed 3D cuboid parameters $d$ to the ground truth $\bm{\hat{d}}$. $p$ is the probability of one pixel in $\bm{m_E}$, and $\bm{\hat{p}}$ is the ground truth of $p$ in $\bm{m_E}$. $q$ is the pixel probability in $\bm{m_C}$, and $\bm{\hat{q}}$ is the ground truth. $n$ is the number of pixels in $\bm{m_E}$ and $\bm{m_C}$ which is the image resolution. 
Note that the RoomNet approach~\cite{lee2017roomnet} uses L2 loss for corner prediction. We discuss the performance using two different losses in Sec.~\ref{text:experiments}. $\alpha, \beta$ and $\tau$ are the weights for each loss term. In our experiment, we set $\alpha = \beta = 1$ and $\tau = 0.01$.

\subsection{Training details}\label{text:training}
Our LayoutNet predicts pixel probabilities for corners and boundaries and regresses the 3D layout parameters.  We find that joint training from a randomly initialized network sometimes fails to converge. Instead, we train each sub-network separately and then jointly train them together. For the 2D layout prediction network, we first train on the layout boundary prediction task to initialize the parameters of the network. For the 3D layout regressor, we first train the network with ground truth layout boundaries and corners as input, and then connect it with the 2D layout decoder and train the whole network end-to-end.

The input Manhattan line map is a 3 channel 0-1 tensor. We normalize each of the 3D cuboid parameter into zero mean and standard deviation across training samples. We use ADAM~\cite{kingma2014_adam} to update network parameters with a learning rate of $e^{-4}$, $\alpha = 0.95$ and $\epsilon = e^{-6}$. The batch size for training the 2D layout prediction network is 5 and changes to 20 for training the 3D regressor. The whole end-to-end training uses a batch size of 20. 

\boldhead{Ground truth smoothing} Our target 2D boundary and corner map is a binary map with a thin curve or point on the 
image. This makes training more difficult. For example, if the network predicts the corner position slightly off the ground truth, a huge penalty will be incurred. Instead, we dilate the ground truth boundary and corner map with a factor of 4 and then smooth the image with a Gaussian kernel of $20\times 20$. 
Note that even after smoothing, the target image still contains \~95\% zero values, so we re-weight the back propagated gradients of the background pixels by multiplying with $0.2$.

\boldhead{Data augmentation}  We use horizontal rotation, left-right flipping and luminance change 
to augment the training samples. The horizontal rotation varies from $0^o-360^o$. The luminance varies with $\gamma$ values between 0.5-2. For perspective images, we apply $\pm 10^\circ$ rotation on the image plane.

\vspace{-0.5em}
\begin{algorithm}
\caption{3D layout optimization}\label{alg}
\begin{algorithmic}[1]
\item Given panorama $I$, layout corner prediction $\bm{m_C}$, and boundary prediction $\bm{m_E}$;
\item Initialize 3D layout $L_0$ based on Eq.~\ref{equ:cuboid};
\State{$E_{best} = Score(L_0)$ by Eq.~\ref{equ:opt}, $L_{best} = L_0$};
\For{$i = 1:$wallNum}
\State{Sample candidate layouts $L_i$ by varying wall position $w_i$ in 3D, fix other wall positions};
\For{$j = 1:|L_i|$}
\State{Sample candidate Layouts $L_{ij}$ by varying floor and ceiling position in 3D};
\State{Rank the best scored Layout $L_B\in \{L_{ij}\}$ based on Eq.~\ref{equ:opt}};
\If{$E_{best} < Score(L_B)$}
\State{$E_{best} = Score(L_{B})$, $L_{best} = L_B$};
\EndIf
\EndFor
\State{Update $w_i$ from $L_{best}$, fix it for following sampling}
\EndFor
\Return{$L_{best}$}
\end{algorithmic}
\end{algorithm}
\vspace{-0.5em}

\subsection{3D layout optimization}\label{text:fitting}
The initial 2D corner predictions are obtained from the corner probability maps that our network outputs. First, the responses are summed across rows, to get a summed response for each column.  Then, local maxima are found in the column responses, with distance between local maxima of at least 20 pixels.  Finally, the two largest peaks are found along the selected columns.  These 2D corners might not satisfy Manhattan constraints, so we perform optimization to refine the estimates. 

Given the predicted corner positions, we can directly recover the camera position and 3D layout, up to a scale and translation, by assuming that bottom corners are on the same ground plane and that the top corners are directly above the bottom ones.  We can further constrain the layout shape to be Manhattan, so that intersecting walls are perpendicular, e.g. like a cuboid or ``L"-shape in a top-down view. For panoramic images, the Manhattan constraints can be easily incorporated, by utilizing the characteristic that the columns of the panorama correspond to rotation angles of the camera. 
We parameterize the layout coordinates in the top-down view as a vector of 2D points $\bm{L_v} = \{ \bm{v_1} = (0, 0), \bm{v_2} = (x_1, y_1), \ldots, \bm{v_N} = (x_N, y_N)\}$.  $\bm{v_1}$ resolves the translation ambiguity, and $|\bm{v_1}-\bm{v_2}| = 1$ sets the scale.  Because the layout is assumed to be Manhattan, neighboring vertices will share one coordinate value, which further reduces the number of free parameters.
We recover the camera position $\bm{v_c}=\{x_c, y_c\}$ and $\bm{L_v}$ based on the following generalized energy minimization inspired by Farin~\etal~\cite{farin2007floor}:
\vspace{-0.5em}
\begin{align}\label{equ:cuboid}
E(\bm{L_v}, \bm{v_c}) = \min_{\bm{v_c}, \bm{L_v}} \sum_{(i,j)\in \bm{L_v}}|\beta(\bm{v_i}, \bm{v_j}) - \alpha(\bm{v_i},\bm{v_j})|
\end{align}
where $\bm{v_i}, \bm{v_j}$ are pairs of neighboring vertices, and $\beta_{ij}=arccos\frac{\bm{v_i}-\bm{v_c}\cdot \bm{v_j}-\bm{v_c}}{\|\bm{v_i}-\bm{v_c}\|\|\bm{v_j}-\bm{v_c}\|}$ is the rotation angle of the camera $\bm{v_c}$ between $\bm{v_i}$ and $\bm{v_j}$. We denote $\alpha_{ij}$ as the pixel-wise horizontal distance on the image between $\bm{v_i}$ and $\bm{v_j}$ divided by the length of the panorama. Note that this $L2$  minimization also applies to general Manhattan layouts. We use L-BFGS~\cite{zhu1997algorithm} to solve for Eq.~\ref{equ:cuboid} efficiently.

We initialize the ceiling level as the average (mean) of 3D upper-corner heights, and then optimize for a better fitting room layout, relying on both corner and boundary information using the following score to evaluate 3D layout candidate $L$:
\vspace{-0.5em}
\begin{align}\label{equ:opt}
Score(L)&= w_{junc}\sum_{l_c\in C}\log P_{\text{corner}}(l_c)\nonumber\\
  &+ w_{ceil}\sum_{l_e\in L_e}\max \log P_{\text{ceil}}(l_e)\nonumber\\
  & + w_{floor}\sum_{l_f\in L_f}\max \log P_{\text{floor}}(l_f)
\end{align}
where $C$ denotes the 2D projected corner positions of $L$. Cardinality of $L$ is \#walls$\times$ 2. We connect the nearby corners on the image to obtain $L_e$ which is the set of projected wall-ceiling boundaries, and $L_f$ which is the set of projected wall-floor boundaries~(each with cardinality of \#walls). $P_{\text{corner}}(\cdot)$ denotes the pixel-wise probability value on the predicted $\bm{m_C}$. $P_{\text{ceil}}(\cdot)$ and $P_{\text{floor}}(\cdot)$ denote the probability on $\bm{m_E}$. The 2nd and 3rd term take the maximum value of log likelihood response in each boundary $l_e\in L_e$ and $l_f\in L_f$. $w_{junc}$, $w_{ceil}$ and $w_{floor}$ are the term weights, we set to 1.0, 0.5 and 1.0 respectively using grid search. This weighting conforms with the observation that wall-floor corners are often occluded, and the predicted boundaries could help improve the layout reconstruction. We find that adding wall-wall boundaries in the scoring function helps less, since the vertical pairs of predicted corners already reveals the wall-wall boundaries information. 

Directly optimizing Eq.~\ref{equ:opt} is computationally expensive, since we penalize on 2D projections but not direct 3D properties. In this case, we instead sample candidate layout shapes and select the best scoring result based on Eq.~\ref{equ:opt}. We use line search to prune the candidate numbers to speed up the optimization. Algorithm~\ref{alg} demonstrates the procedure. 
In each step, we sample candidate layouts by shifting one of the wall position within $\pm \%10$ of its distance to the camera center. Each candidate's ceiling and floor level is then optimized based on the same sampling strategy and scored based on Eq.~\ref{equ:opt}. Once we find the best scored layout by moving one of the walls, we fix this wall position, move to the next wall and perform the sampling again. We start from the least confident wall based on our boundary predictions. In total, $\sim 1000$ layout candidates are sampled. The optimization step spends less then 30 sec for each image and produces better 3D layouts as demonstrated in Sec.~\ref{text:experiments}. 

\subsection{Extensions}\label{text:exten}

With small modifications, our network, originally designed to predict cuboid layouts from panoramas, can also predict more general Manhattan layouts from panoramas and cuboid-layouts from perspective images.

\boldhead{General Manhattan layouts} To enable more general layouts, we include training examples that have more than four walls visible (e.g. ``L''-shaped rooms), which applies to about $10\%$ of examples. We then determine whether to generate four or six walls by thresholding the score of the sixth strongest wall-wall boundary. Specifically, the average probability along the sixth strongest column of the corner map is at least 0.05. In other words, if there is evidence for more than four walls, our system generates additional walls; otherwise it generates four. Since the available test sets do not have many examples with more than four walls, we show qualitative results with our additional captured samples in Sec.~\ref{text:experiments_noncuboid} and in the supplemental material.

Note that there will be multiple solutions given non-cuboid layout when solving Eq.~\ref{equ:cuboid}. We experimented with predicting a concave/convex label as part of the corner map prediction to obtain single solution, but observed degraded 2D prediction. We thus enumerate all possible shapes (e.g. for room with six walls, there will be six variations) and choose the one with the best score. We found this heuristic search to be efficient as it searches in a small discrete set. We do not train with the 3D parameter regressor for the non-cuboid layout. 

\boldhead{Perspective images}
When predicting on perspective images, we skip the alignment and optimization steps, instead directly predicting corners and boundaries on the image. We also do not use the 3D regressor branch. The network predicts a 3-channel boundary layout map with ceiling-wall, wall-wall and wall-floor boundaries, and the corner map has eight channels for each possible corner. Since perspective images have smaller fields of view and the number of visible corners varies, we add a small decoding branch that predicts the room layout type, similar to RoomNet~\cite{lee2017roomnet}. The predictor has 4 fully-connected (fc) layers with 1024, 256, 64 and 11 nodes, with ReLU operations in between. The predicted layout type then determines which corners are detected, and the corners are localized as the most probable positions in the corner maps. We use cross entropy loss to jointly train the layout boundary and corner predictors. To ease training, similar to the procedure in Sec.~\ref{text:training}, we first train the boundary/corner predictors, and then add the type predictor branch and train all components together.


\section{Experiments}\label{text:experiments}
We implement our LayoutNet with Torch and test on a single NVIDIA Titan X GPU. The layout optimization is implemented with Matlab R2015a and is performed on Linux machine with Intel Xeon 3.5G Hz in CPU mode. 

We demonstrate the effectiveness of our approach on the following tasks: 1) predict 3D cuboid layout from a single panorama, 2) estimate 3D non-cuboid Manhattan layout from a single panorama, and 3) estimate layout from a single perspective image. We train only on the training split of each public dataset and tune the hyper-parameters on the validation set. We report results on the test set. Our final corner/boundary prediction from the LayoutNet is averaged over results with input of the original panoramas/images and the left-right flipped ones. Please find more results in the supplemental materials.

\begin{figure*}[ht]
\begin{center}
\includegraphics[width=0.94\linewidth]{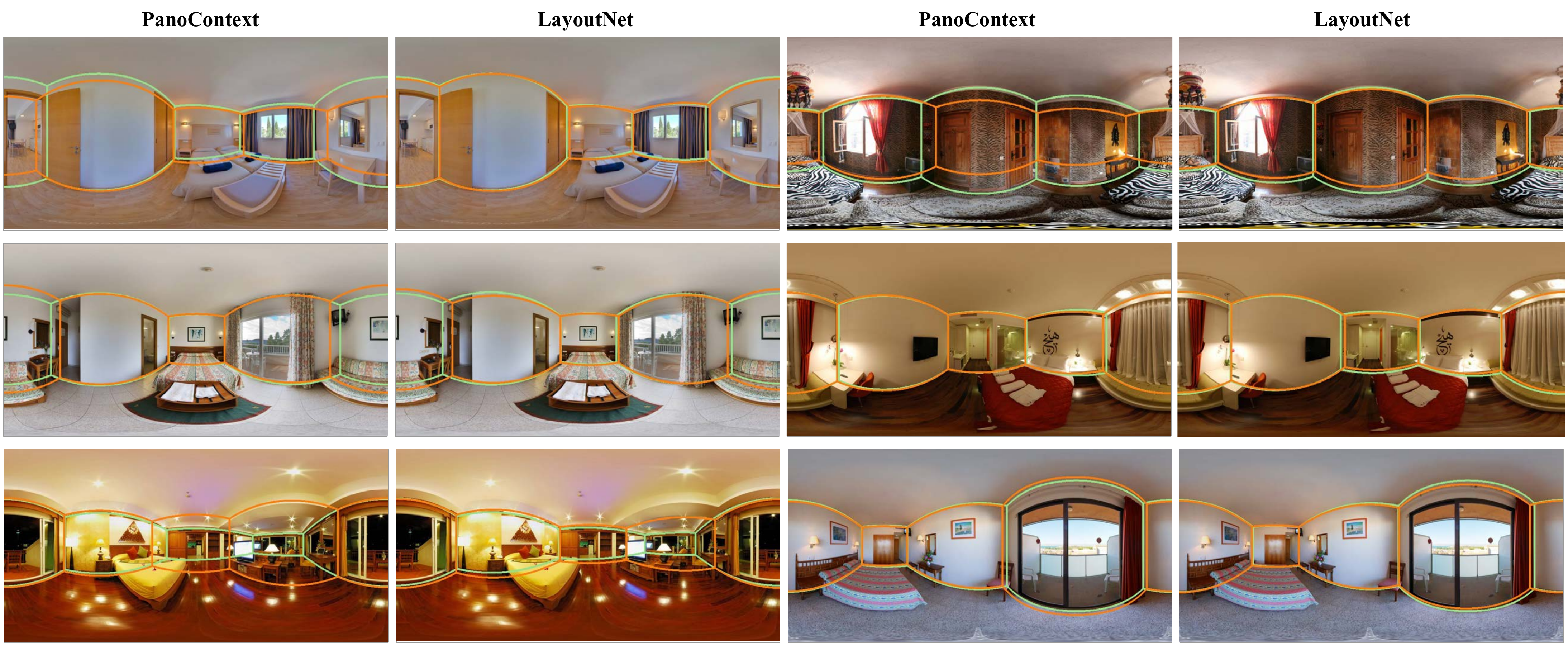}
\end{center}
\vspace{-2.5mm}
   \caption{\textbf{Qualitative results~(randomly sampled) for cuboid layout prediction on PanoContext dataset~\cite{zhang2014panocontext}}. We show both our method's performance~(even columns) and the state-of-the-art~\cite{zhang2014panocontext}~(odd columns). Each image consists predicted layout from given method~(orange lines) and ground truth layout~(green lines). Our method is very accurate on the pixel level, but as the IoU measure shows in our quantitative results, the 3D layout can be sensitive to even small 2D prediction errors. Best viewed in color. }
   \vspace{-1.0em}
\label{fig:cuboid}
\end{figure*}

\begin{figure*}[h]
\begin{center}
\includegraphics[width=0.94\linewidth]{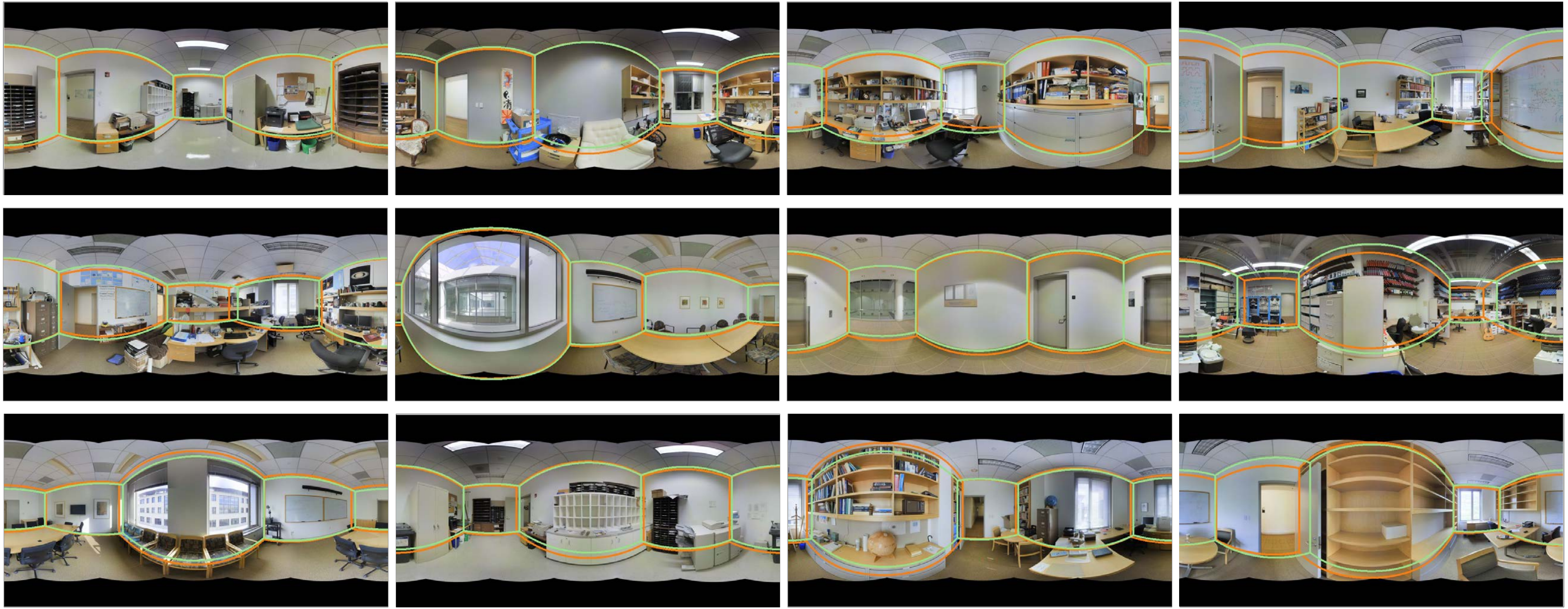}
\end{center}
\vspace{-2.5mm}
   \caption{\textbf{Qualitative results~(randomly sampled) for cuboid layout prediction on the Stanford 2D-3D annotation dataset.} This dataset is more challenging than the PanoContext dataset, due to a smaller vertical field of view and more occlusion. We show our method's predicted layout~(orange lines) compared with the ground truth layout~(green lines). 
   Best viewed in color.}
   \vspace{-1.0em}
\label{fig:stanford}
\end{figure*}

\begin{figure*}[h]
\begin{center}
\includegraphics[width=0.86\linewidth]{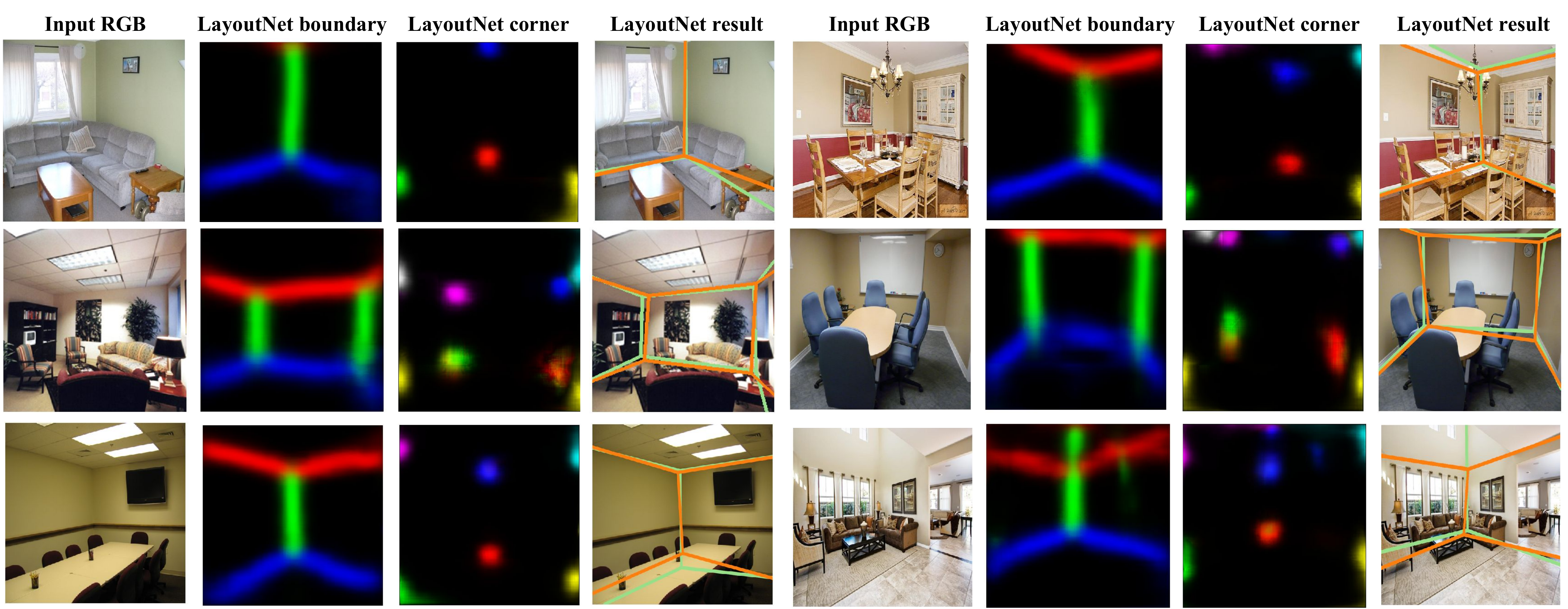}
\end{center}
\vspace{-2.5mm}
   \caption{\textbf{Qualitative results for perspective images}. We show the input RGB image, our predicted boundary/corner map and the final estimated layout (orange lines) compared with ground truth (green lines). Best viewed in color.}
   \vspace{-1.5em}
\label{fig:persp}
\end{figure*}

\begin{figure}[ht]
\begin{center}
\includegraphics[width=0.95\linewidth]{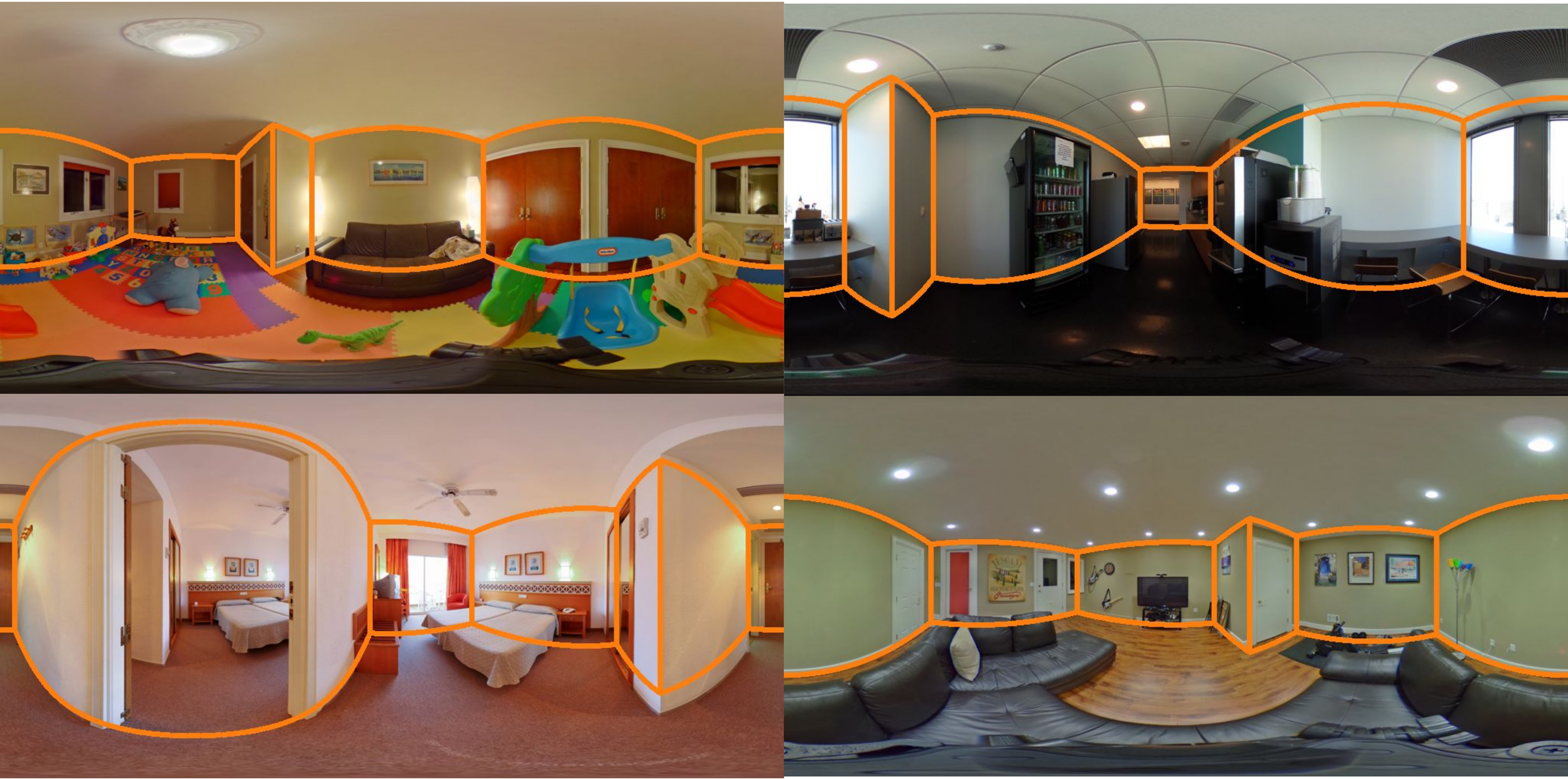}
\end{center}
\vspace{-2mm}
   \caption{\textbf{Qualitative results for non-cuboid layout prediction.} We show our method's predicted layout (orange lines) for non-cuboid layouts such as ``L"-shaped rooms.  Best viewed in color. }
   \vspace{-1.5em}
\label{fig:noncuboid}
\end{figure}

\subsection{Cuboid layout for panorama}


\begin{table}
\begin{center}
\resizebox{0.5\textwidth}{!}{
\begin{tabular}{|c|c|c|c|c|}
\hline
Method & 3D IoU~(\%) & \begin{tabular}{c} Corner \\ error~(\%)  \end{tabular}  & \begin{tabular}{c} Pixel \\ error~(\%)  \end{tabular}\\
\hline\hline
PanoContext~\cite{zhang2014panocontext}&67.23 & 1.60 & 4.55\\
\hline
ours~(corner)&73.16 &1.08&4.10\\
ours~(corner+boundary)& 73.26& 1.07& \textbf{3.31}\\
ours full~(corner+boundary+3D)&\textbf{74.48}& \textbf{1.06}& 3.34\\
\hline
ours w/o alignment& 69.91 & 1.44 & 4.39\\
ours w/o cuboid constraint & 72.56& 1.12&3.39\\
ours w/o layout optimization &73.25& 1.08 & 3.37\\
ours w/ $L2$ loss& 73.55 & 1.12 & 3.43\\
\hline
ours full w/ Stnfd. 2D-3D data  & 75.12& 1.02 & 3.18\\
\hline
\end{tabular}
}
\end{center}
\vspace{-1.5mm}
\caption{Quantitative results on cuboid layout estimation from panorama using PanoContext dataset~\cite{zhang2014panocontext}. We compare the PanoContext method, and include an ablation analysis on a variety of configurations of our method. Bold numbers indicate the best performance when training on PanoContext data.}
\vspace{-0.5em}
\label{tab:cuboid}
\end{table}

\begin{table}
\begin{center}
\resizebox{0.4\textwidth}{!}{
\begin{tabular}{|c|c|}
\hline
Method & Average CPU time (s)\\
\hline
PanoContext~\cite{zhang2014panocontext} & $>$ 300\\
ours full~(corner+boundary+3D)& 44.73\\
\hline
ours w/o alignment &~31.00\\ 
ours w/o cuboid constraint &13.75\\ 
ours w/o layout optimization &14.23\\
\hline
\end{tabular}
}
\end{center}
\vspace{-1.5mm}
\caption{Average CPU time for each method. We evaluate the methods on the PanoContext dataset~\cite{zhang2014panocontext} using Matlab on Linux machine with an Intel Xeon 3.5G Hz (6 cores).}
\vspace{-1.0em}
\label{tab:time}
\end{table}

\begin{table}
\begin{center}
\resizebox{0.5\textwidth}{!}{
\begin{tabular}{|c|c|c|c|}
\hline
Method & 3D IoU~(\%)& \begin{tabular}{c} Corner \\ error~(\%)  \end{tabular}  & \begin{tabular}{c} Pixel \\ error~(\%)  \end{tabular}\\
\hline
ours~(corner)& 72.50 &1.27 & 3.44\\
ours~(corner+boundary) & 75.26&1.03 & \textbf{2.68}\\
ours full~(corner+boundary+3D)& 75.39& \textbf{1.01} & 2.70\\
\hline
ours w/o alignment& 68.56 & 1.56& 3.70\\
ours w/o cuboid constraint &74.13&1.08 &2.87\\
ours w/o layout optimization&74.47 &1.07&2.92\\
ours w/ $L2$ loss & \textbf{76.33}& 1.04 & 2.70\\
\hline
ours full w/ PanoContext data & 77.51& 0.92 & 2.42\\
\hline
\end{tabular}
}
\end{center}
\vspace{-1.5mm}
\caption{Evaluation on our labeled Stanford 2D-3D annotation dataset. We evaluate our LayoutNet approach with various configurations for ablation study.  Bold numbers indicate best performance when training only on Stanford 2D-3D training set.}
\vspace{-1.0em}
\label{tab:cuboid_stan}
\end{table}

We evaluate our approach on three standard metrics: 
\begin{enumerate}
\vspace{-0.5em}
    \item 3D Intersection over Union (IoU), calculated between our predicted 3D layout and the ground truth and averaged across all images;
    \vspace{-0.5em}
    \item Corner error, the $L2$ distance between predicted room corner and the ground truth, normalized by the image diagonal and averaged across all images;
    \vspace{-0.5em}
    \item Pixel error, the pixel-wise accuracy between the layout and the ground truth, averaged across all images.
    \vspace{-0.5em}
\end{enumerate}

We perform our method using the same hyper-parameter on the following two datasets.

\boldhead{PanoContext dataset} The PanoContext dataset~\cite{zhang2014panocontext} contains $~500$ annotated cuboid layouts of indoor environments such as bedrooms and living rooms. Since there is no existing validation set, we carefully split 10\% validation images from the training samples so that similar rooms do not appear in the training split. Table~\ref{tab:cuboid} shows the quantitative comparison of our method, denoted as ``ours full~(corner+boundary+3D)", compared with the state-of-the-art cuboid layout estimation by Zhang~\etal~\cite{zhang2014panocontext}, denoted as ``PanoContext". Note that PanoContext incorporates object detection as a factor for layout estimation. Our LayoutNet directly recovers layouts and outperforms the state-of-the-art on all the three metrics. Figure~\ref{fig:cuboid} shows the qualitative comparison. Our approach presents better localization of layout boundaries, especially for a better estimate on occluded boundaries, and is much faster in time as shown in Table~\ref{tab:time}.

\boldhead{Our labeled Stanford 2D-3D annotation dataset} The dataset contains 1413 equirectangular RGB panorama collected in 6 large-scale indoor environment including office and classrooms and open space like corridors. Since the dataset does not contain applicable layout annotations, we extend the annotations with carefully labeled 3D cuboid shape layout, providing $571$ RGB panoramas with room layout annotations. We evaluate our LayoutNet quantitatively in Table~\ref{tab:cuboid_stan} and qualitatively in Figure~\ref{fig:stanford}. Although the Stanford 2D-3D annotation dataset is more challenging with smaller vertical field of view (FOV) and more occlusions on the wall-floor boundaries, our LayoutNet recovers the 3D layouts well. 

\boldhead{Ablation study} We show, in Table~\ref{tab:cuboid} and Table~\ref{tab:cuboid_stan}, the performance given the different configurations of our approach: 1) with only room corner prediction, denoted as ``ours~(corner)''; 2) joint prediction of corner and boundary, denoted as ``ours~(corner+boundary)''; 3) our full approach with 3D layout loss, denoted as ``ours full~(corner+boundary+3D)''; 4) our full approach trained on a combined dataset; 5) our full approach without alignment step; 6) our full approach without cuboid constraint; 7) our full approach without layout optimization step; and 8) our full approach using $L2$ loss for boundary/corner prediction instead of cross entropy loss. Our experiments show that the full approach that incorporates all configurations performs better across all the metrics. 
Using cross entropy loss appears to have a better performance than using L2. Training with 3D regressor has a small impact, which is the part of the reason we do not use it for perspective images. Table~\ref{tab:time} shows the average runtimes for different configurations.

\begin{table}
\begin{center}
\resizebox{0.31\textwidth}{!}{
\begin{tabular}{c|c|c}
\hline
Method & L2 dist & cosine dist\\
\hline
Yang et al.~\cite{yang2016efficient} & 27.02 & \textbf{4.27}\\
Ours & \textbf{18.51}& 5.85\\
\hline
\end{tabular}
}
\end{center}
\vspace{-1mm}
\caption{Depth distribution error compared with Yang~\etal~\cite{yang2016efficient}.} 
\vspace{-2.0em}
\label{tab:pano_comp}
\end{table}

\textbf{Comparison to other approaches:} We compare with Yang~\etal{} based on their depth distribution metric. We directly run our full cuboid layout prediction (deep net trained on PanoContext + optimization) on 88 indoor panoramas collected by Yang~\etal{} 
As shown in Table~\ref{tab:pano_comp}, our approach outperforms Yang~\etal{} in L2 distance and is slightly worse in cosine distance. Another approach, Pano2CAD~\cite{Pano2CAD}, has not made their source code available and has no evaluation on layout, making direct comparison difficult. For time consumption, Yang~\etal{} report to be less than 1 minute, Pano2CAD takes 30s to process one room. One forward pass of LayoutNet takes 39ms. In CPU mode~(w/o parallel for loop) using Matlab R2015a, our cuboid constraint takes 0.52s, alignment 13.73s, and layout optimization 30.5s.

\subsection{Non-cuboid layout for panorama}\label{text:experiments_noncuboid}
Figure~\ref{fig:noncuboid} shows qualitative results of our approach to reconstruct non-cuboid Manhattan layouts from single panorama. Due to the limited number of non-cuboid room layouts in the existing datasets, we captured several images using a Ricoh Theta-S 360$^\circ$ camera. Our approach is able to predict 3D room layouts with complex shape that are difficult for existing methods.

\subsection{Perspective images}
We use the same experimental setting as in~\cite{dasgupta2016delay, lee2017roomnet}. We train our modified approach to jointly predict room type on the training split of the LSUN layout estimation challenge. We do not train on the validation split. 

Table~\ref{tab:persp} shows our performance compared with the state-of-the-art on Hedau's dataset~\cite{hedau2009recovering}. Our method ranks second among the methods. Our method takes 39ms (25 FPS) to process a perspective image, faster than the 52ms (19 FPS) of RoomNet basic~\cite{lee2017roomnet} or 168ms (6 FPS) of RoomNet recurrent, under the same hardware configuration. 
We report the result on LSUN dataset in the supplemental material. 
Figure~\ref{fig:persp} shows qualitative results on the LSUN validation split. Failure cases include room type prediction error~(last row, right column) and heavy occlusion from limited field of view~(last row, left column).

\begin{table}
\begin{center}
\resizebox{0.35\textwidth}{!}{
\begin{tabular}{c|c}
\hline
Method & Pixel Error~(\%)\\
\hline
Schwing et al.~\cite{schwing2012efficient_cvpr} & 12.8\\
Del Pero et al.~\cite{del2013understanding}& 12.7\\
Dasgupta et al.~\cite{dasgupta2016delay}  & 9.73\\ 
LayoutNet~(ours)  & 9.69\\ 
RoomNet recurrent 3-iter~\cite{lee2017roomnet} & \textbf{8.36}\\
\hline
\end{tabular}
}
\end{center}
\vspace{-1mm}
\caption{Performance on Hedau dataset~\cite{hedau2009recovering}. We show the top 5 results, LayoutNet ranks second to RoomNet recurrent 3-iter in Pixel Error~(\%).} 
\vspace{-1.5em}
\label{tab:persp}
\end{table}

\section{Conclusion}
We propose LayoutNet, an algorithm that predicts room layout from a single panorama or perspective image. Our approach relaxes the commonly assumed cuboid layout limitation and works well with non-cuboid layouts (e.g. ``L"-shape room). We demonstrate how pre-aligning based on vanishing points and Manhattan constraints substantially improve the quantitative results. Our method operates directly on panoramic images (rather than decomposing into perspective images) and is among the state-of-the-art for the perspective image task. Future work includes extending to handle arbitrary room layouts, incorporating object detection for better estimating room shapes, and recovering a complete 3D indoor model recovered from single images.


\section*{Acknowledgements}
\vspace{-1.5mm}
This research is supported in part by NSF award 14-21521, ONR MURI grant N00014-16-1-2007, and Zillow Group. We thank Zongyi Wang for his invaluable help with panorama annotation.


{\small
\bibliographystyle{ieee}
\bibliography{egbib}
}
\newpage
\appendix 
\section{Quantitative Results on LSUN layout Challenge~\cite{lsun_challenge}}

Table~\ref{tab:persp_supp} shows our performance compared with the state-of-the-art on the LSUN dataset~\cite{lsun_challenge}. Our method ranks second in Keypoint Error~(\%) and ranks third in Pixel Error~(\%) among the methods. We also report results of the RoomNet basic approach~\cite{lee2017roomnet} that does not apply recurrent refinement, which is closer in design to our approach.

The lower accuracy in pixel error mainly results from our simplified room keypoint representation. Different from RoomNet~\cite{lee2017roomnet} that assumes all keypoints are distinguished across different room types, our LayoutNet directly predicts the 8 keypoints, and selects among them based on the room type to produce the final prediction. Applying the layout optimization step as explained in the paper could possibly further enhance our performance on the perspective image task.


\begin{table}[h]
\begin{center}
\resizebox{0.5\textwidth}{!}{
\begin{tabular}{c|c|c}
\hline
Method & Keypoint Error~(\%) & Pixel Error~(\%) \\
\hline
Hedau \etal~\cite{hedau2009recovering} & 15.48 & 24.23\\
Mallya \etal~\cite{mallya2015learning}& 11.02 & 16.71\\
Dasgupta \etal~\cite{dasgupta2016delay}  & 8.20 & 10.63\\ 
LayoutNet~(ours)  & 7.63 & 11.96\\ 
RoomNet recurrent 3-iter~\cite{lee2017roomnet} & \textbf{6.30} &
\textbf{9.86}\\
\hline
RoomNet basic~\cite{lee2017roomnet} &6.95 &10.46\\
\hline
\end{tabular}
}
\end{center}
\caption{Performance on LSUN dataset~\cite{lsun_challenge}. LayoutNet ranks second to RoomNet recurrent 3-iter in Keypoint Error~(\%) and ranks third in Pixel Error~(\%). We also report the RoomNet basic approach that does not apply recurrent refinement step.} 
\label{tab:persp_supp}
\end{table}

\section{More Qualitative Results}

\subsection{Non-cuboid layout from panorama}
\begin{figure*}[ht]
\begin{center}
\includegraphics[width=0.99\linewidth]{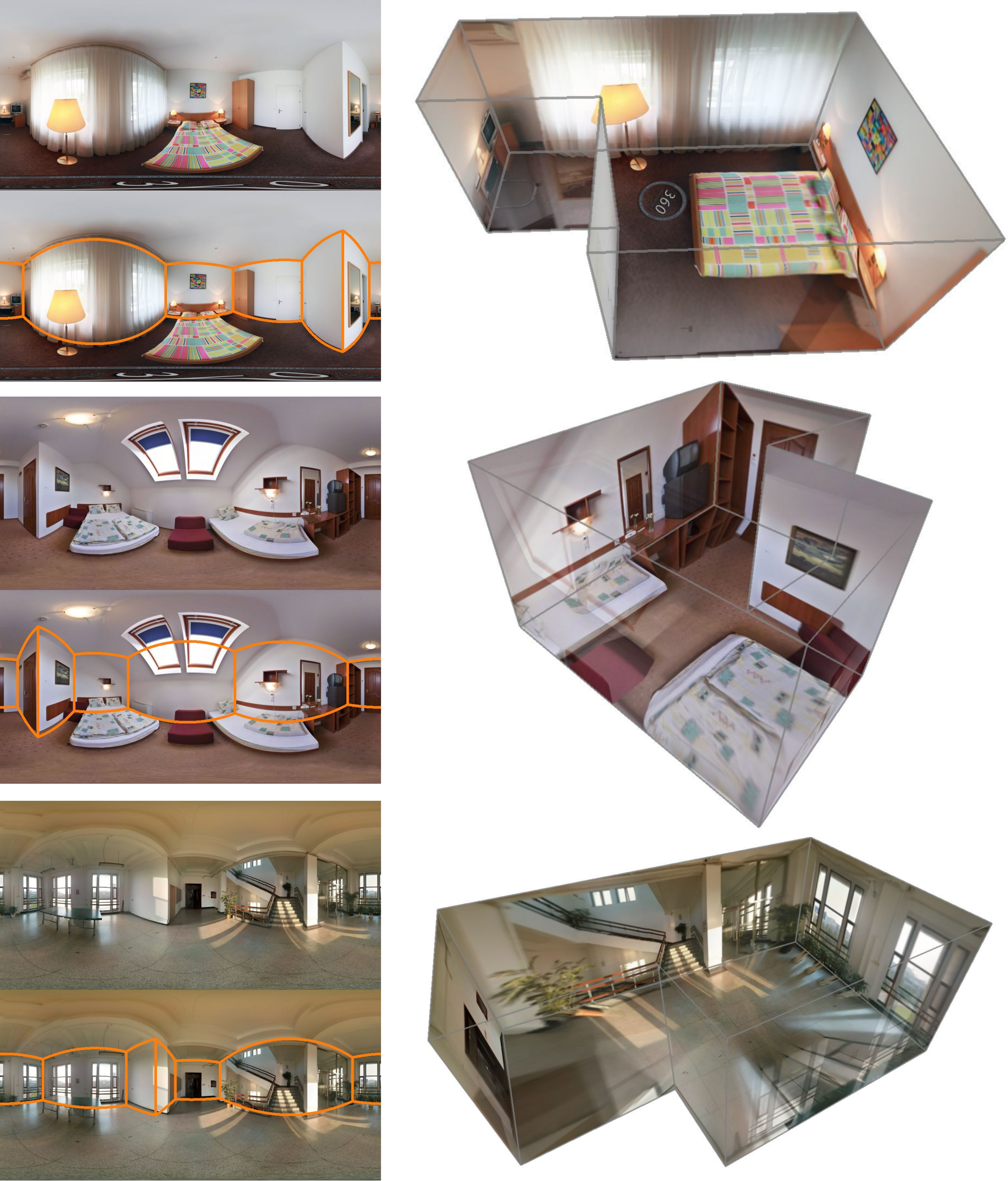}
\end{center}
   \caption{\textbf{Qualitative results for non-cuboid layout prediction.} We show our method's predicted layout (orange lines) for non-cuboid layouts such as ``L"-shaped rooms.  Best viewed in color.}
\label{fig:noncuboid_1}
\end{figure*}

We show more qualitative results of non-cuboid room layout reconstruction from single panorama as in Figure~\ref{fig:noncuboid_1}. We use samples from the dataset collected by Yang~\etal~\cite{yang2016efficient}. We exclude samples that overlap with the PanoContext dataset~\cite{zhang2014panocontext}.

\subsection{Cuboid layout from panorama}

\begin{figure*}[ht]
\begin{center}
\includegraphics[width=0.99\linewidth]{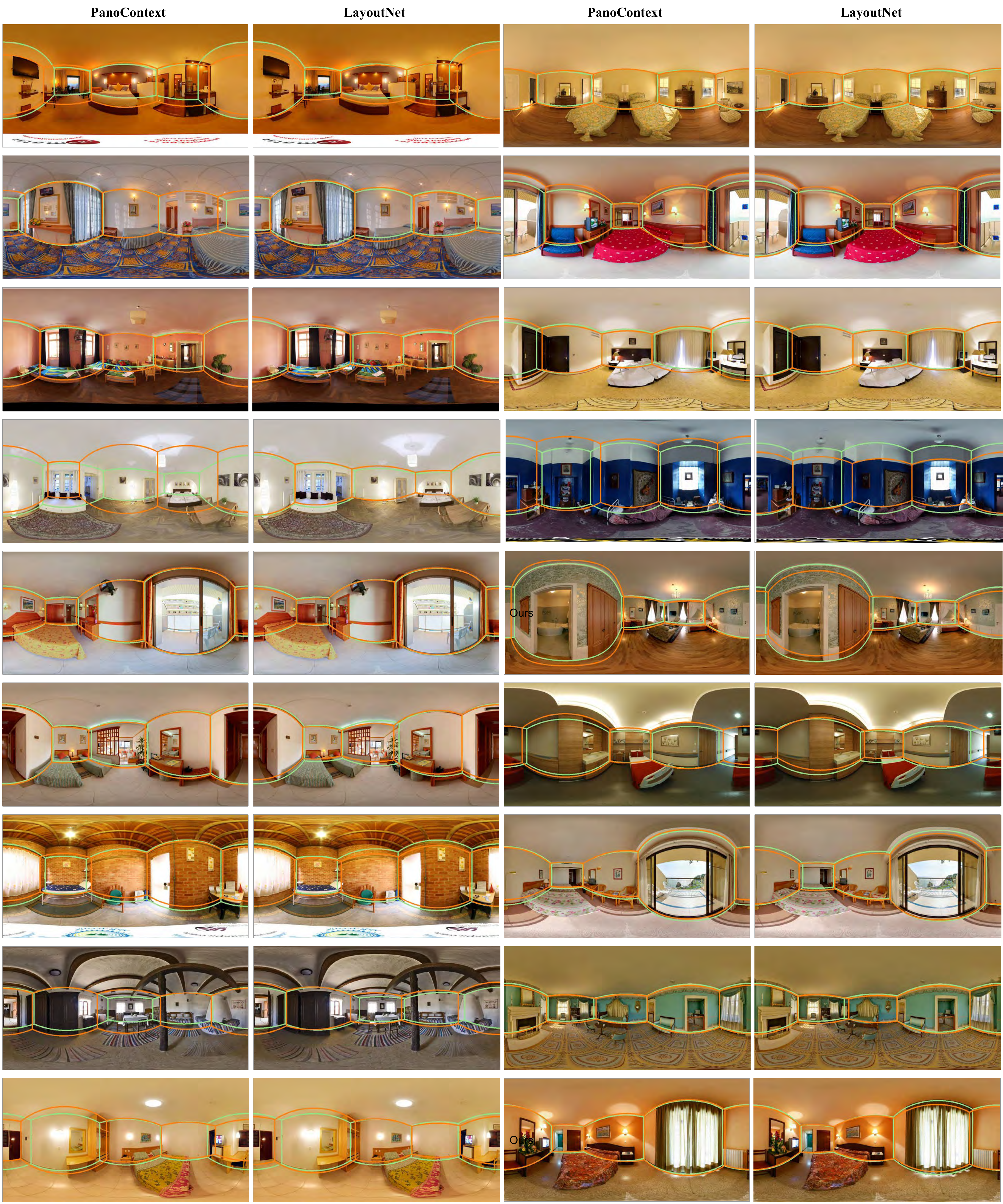}
\end{center}
   \caption{\textbf{Qualitative results for cuboid layout prediction on PanoContext dataset~\cite{zhang2014panocontext}}. We show both our method's performance~(even columns) and the state-of-the-art~\cite{zhang2014panocontext}~(odd columns). Each image consists predicted layout from given method~(orange lines) and ground truth layout~(green lines). Best viewed in color. }
\label{fig:cuboid_1}
\end{figure*}

\begin{figure*}[ht]
\begin{center}
\includegraphics[width=0.99\linewidth]{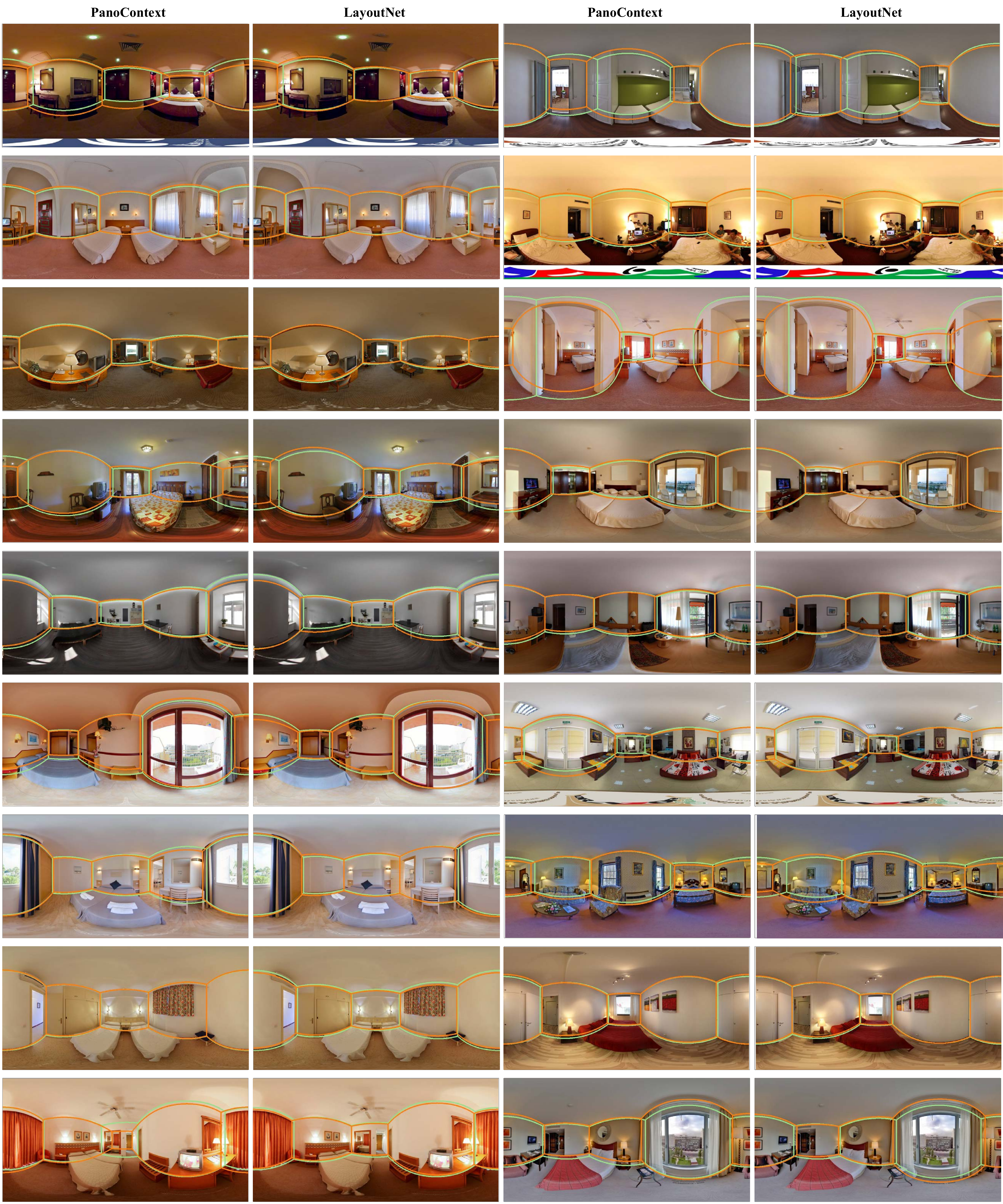}
\end{center}
   \caption{\textbf{Qualitative results for cuboid layout prediction on PanoContext dataset~\cite{zhang2014panocontext}}. We show both our method's performance~(even columns) and the state-of-the-art~\cite{zhang2014panocontext}~(odd columns). Each image consists predicted layout from given method~(orange lines) and ground truth layout~(green lines). Best viewed in color. }
\label{fig:cuboid_2}
\end{figure*}

We show more qualitative results in PanoContext dataset~\cite{zhang2014panocontext} in Figure~\ref{fig:cuboid_1} and Figure~\ref{fig:cuboid_2}. We compare our method with the state-of-the-art.

\begin{figure*}[ht]
\begin{center}
\includegraphics[width=0.99\linewidth]{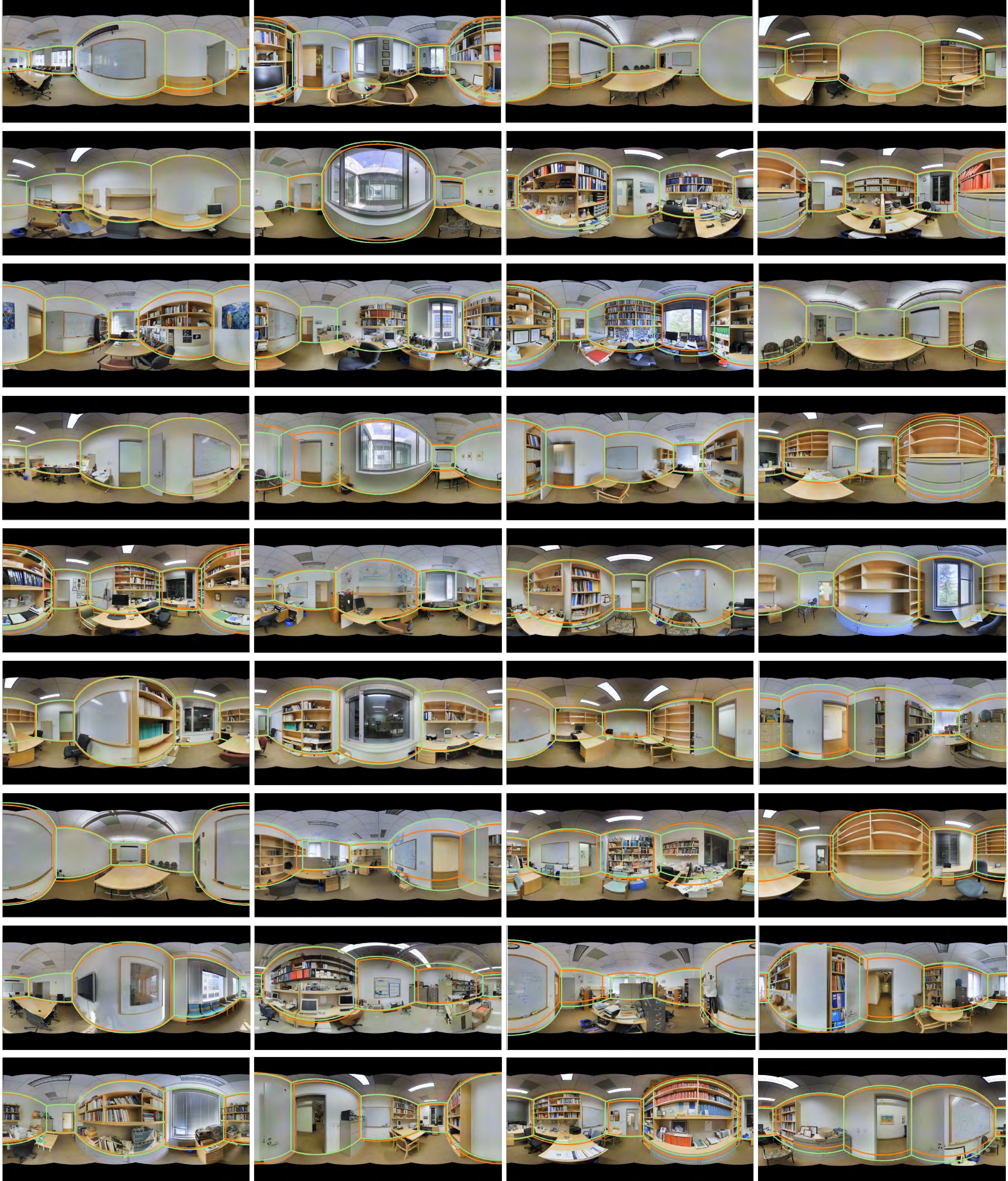}
\end{center}
   \caption{\textbf{Qualitative results~(randomly sampled) for cuboid layout prediction on the Stanford 2D-3D annotation dataset.} This dataset is more challenging than the PanoContext dataset, due to a smaller vertical field of view and more occlusion. We show our method's predicted layout~(orange lines) compared with the ground truth layout~(green lines).
   Best viewed in color.}
\label{fig:stanford_1}
\end{figure*}

\begin{figure*}[ht]
\begin{center}
\includegraphics[width=0.99\linewidth]{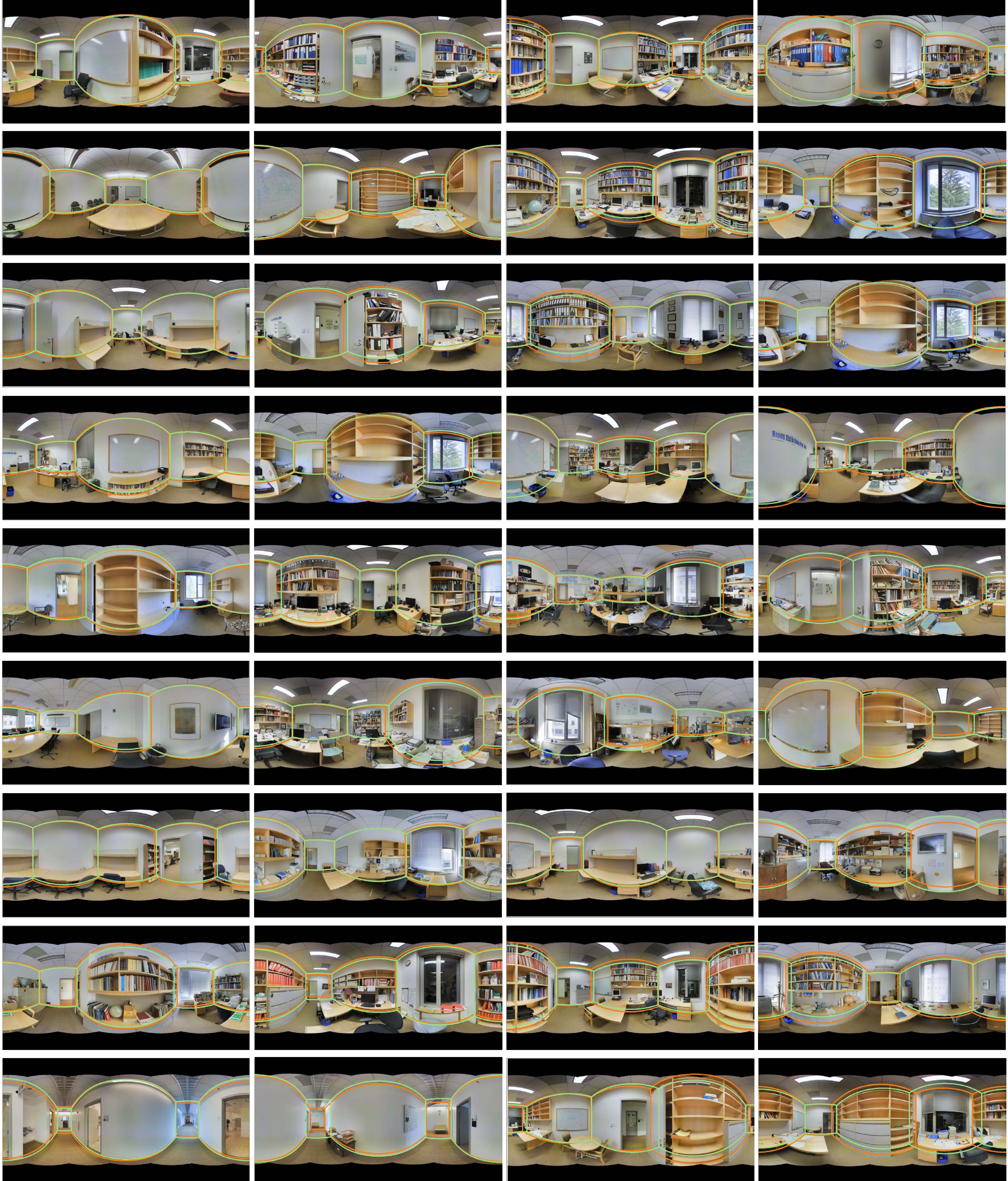}
\end{center}
   \caption{\textbf{Qualitative results~(randomly sampled) for cuboid layout prediction on the Stanford 2D-3D annotation dataset.} This dataset is more challenging than the PanoContext dataset, due to a smaller vertical field of view and more occlusion. We show our method's predicted layout~(orange lines) compared with the ground truth layout~(green lines).
   Best viewed in color.}
\label{fig:stanford_2}
\end{figure*}

We show more qualitative results in our labeled Stanford 2D-3D annotation dataset compared with our ground truth annotation, as shown in Figure~\ref{fig:stanford_1} and Figure~\ref{fig:stanford_2}.

\subsection{Perspective images}

\begin{figure*}[ht]
\begin{center}
\includegraphics[width=0.86\linewidth]{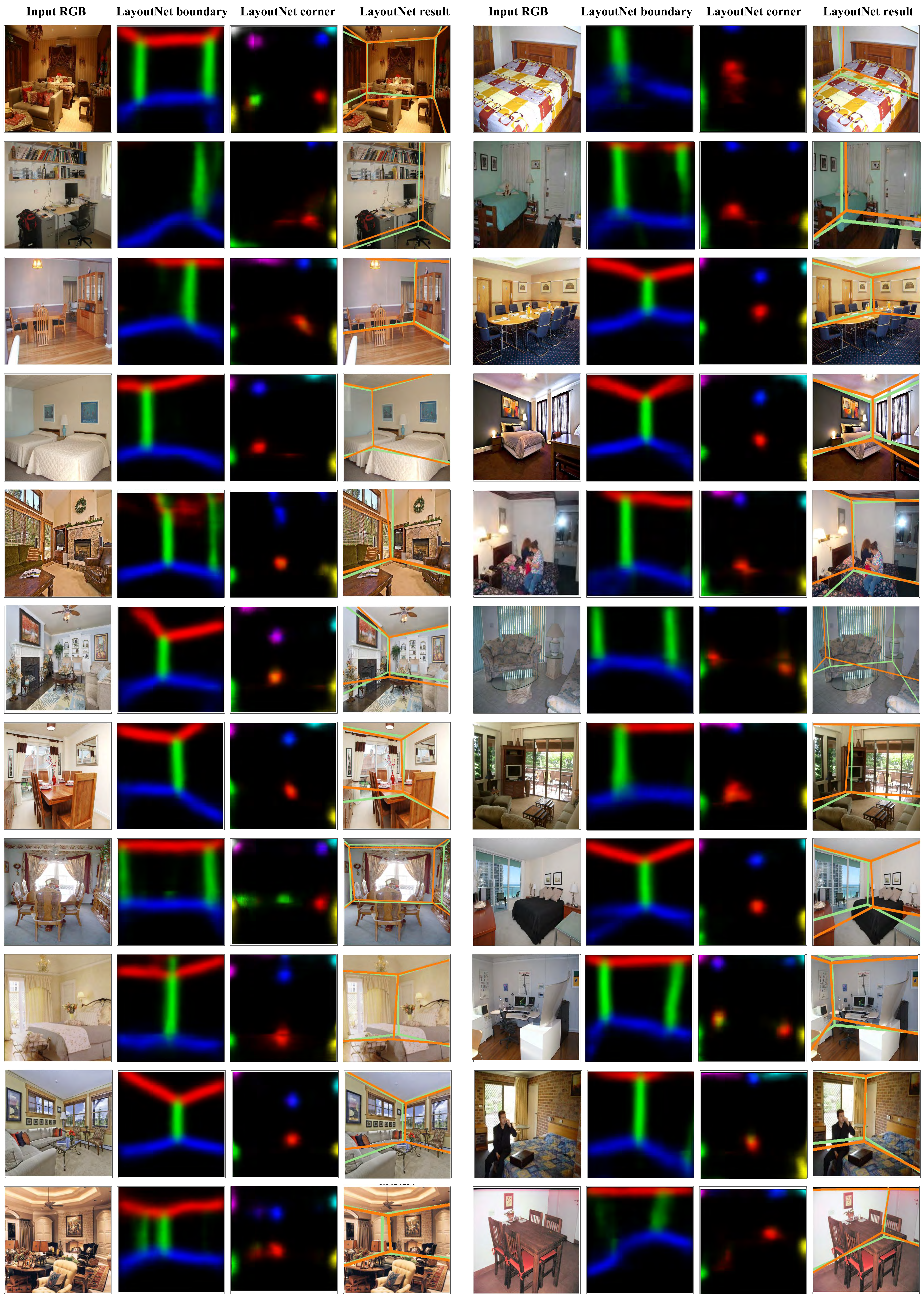}
\end{center}
   \caption{\textbf{Qualitative results for perspective images}. We show the input RGB image, our predicted boundary/corner map and the final estimated layout (orange lines) compared with ground truth (green lines). Best viewed in color.}
\label{fig:lsun_1}
\end{figure*}

We show more qualitative results on the LSUN layout Challenge~\cite{lsun_challenge} compared with the ground truth annotation, as shown in Figure~\ref{fig:lsun_1}.

\end{document}